\documentclass[10pt,letterpaper,compsoc,conference]{iiswc22}

\usepackage{cite}
\usepackage{amsmath,amssymb,amsfonts}
\usepackage{algorithmic}
\usepackage{graphicx}
\usepackage[dvipsnames]{xcolor}
\usepackage[final]{microtype}
\usepackage[italic]{mathastext}
\usepackage{libertine}
\usepackage[T1]{fontenc}
\usepackage{textcomp}
\usepackage[varqu,varl]{zi4}
\usepackage[all]{nowidow}
\usepackage[auth-lg,affil-it]{authblk}
\usepackage[keeplastbox]{flushend}

\usepackage[bookmarks=true,breaklinks=true,letterpaper=true,colorlinks,linkcolor=black,citecolor=blue,urlcolor=black]{hyperref}
\usepackage{algorithm}
\usepackage{comment}
\usepackage{multirow}
\usepackage{enumitem}
\usepackage{wrapfig}
\usepackage{grffile}

\begin{document}

%% EDIT TITLE BELOW

\title{FedGPO: Heterogeneity-Aware Global Parameter Optimization for Efficient Federated Learning
%FedGPO: Characterizing and Designing for \\ 
%Efficient Federated Learning using \\ Heterogeneity-Aware Global Parameter Optimization
% FedGPO: Heterogeneity-Aware Global Parameter Optimization for Federated Learning
\vspace{-0.3cm}
}

%% DO NOT EDIT THE FOLLOWING

\renewcommand\Authsep{\qquad}
\renewcommand\Authand{\qquad}
\renewcommand\Authands{\qquad}

%% EDIT AUTHOR LIST BELOW

\begin{comment}
\author{Author1 Name}
\author{Author2 Name}
\author{Author3 Name}
\affil{Full Name of Awesome School}
\end{comment}

\author{Young Geun Kim \;\;\;\;\;\;\;\;\;\; Carole-Jean Wu\\ 
\normalsize \;\;\;\;\; Korea University \;\;\;\;\;\;\;\;\;\;\;\;\;\;\;\;\;\;\;\;\; ASU / Meta \;\;\;\;\;\;\;\; \\
\normalsize \;\;younggeun\_kim@korea.ac.kr \;\;\;\; carole-jean.wu@asu.edu \;\;\;\;\;
\vspace{-0.3cm}
}

%%% ALTERNATIVE FORMAT FOR MULTIPLE SCHOOLS:
%%% 
% \author[1]{Author1 Name}
% \author[2]{Author2 Name}
% \author[2]{Author3 Name}
% \author[1]{Author4 Name}
% \affil[1]{Full Name of Awesome School}
% \affil[2]{Full Name of Awesomer School}

\maketitle

%% EDIT YOUR PAPER'S CONTENTS BELOW

\begin{abstract}
\vspace{-0.1cm}
Federated learning (FL) has emerged as a solution to deal with the risk of privacy leaks in machine learning training. This approach allows a variety of mobile devices to collaboratively train a machine learning model without sharing the raw on-device training data with the cloud. However, efficient edge deployment of FL is challenging because of the system/data heterogeneity and runtime variance. This paper optimizes the energy-efficiency of FL use cases while guaranteeing model convergence, by accounting for the aforementioned challenges. We propose FedGPO based on a reinforcement learning, which learns how to identify optimal global parameters (\textit{B, E, K}) for each FL aggregation round adapting to the system/data heterogeneity and stochastic runtime variance. In our experiments, FedGPO improves the model convergence time by 2.4 times, and achieves 3.6 times higher energy efficiency over the baseline settings, respectively.
\end{abstract}
\section{Introduction}
\label{sec:introduction}

Federated learning (FL) has recently emerged as a practical framework for training machine learning (ML) models on a large variety of mobile devices privately~\cite{KBonawitz2019,JKonecny2016,HBMcMahan2017,Dzmitry:mlsys-papaya}. A shared ML model is trained over \textit{E} epochs with a minibatch size of \textit{B} on \textit{K} selected devices, where \textit{K} is a subset of \textit{N} client devices participating in the FL task. The \textit{K} devices then upload the respective model gradients (or trained model parameters) to the cloud, in order to update the global model while keeping all the raw data on the device. Thus, FL deals with the risk of privacy leaks for deep neural network (DNN) training.

While many existing approaches have been proposed to deploy FL efficiently~\cite{HBMcMahan2017,YDeng2021,CChen2021,TLi2020,YGKim2021,JKonecny2016,YJin2020,CWang2021}, a fundamental challenge remains --- setting global parameters (\textit{B, E, K}) round-by-round to ensure efficient edge execution. These global parameters significantly affect the model quality and convergence time, as they directly determine the amount of data reflected on the model gradients~\cite{MKhodak2021}. Moreover, they also affect the energy efficiency of participant devices, because the amount of training computation on each client device depends on the parameter settings~\cite{YGKim2021}. Therefore, to achieve efficient FL execution at the edge, finding optimal global parameter settings is crucial. 

Hyperparameter optimization (HPO) has been extensively studied for the centralized training. Common approaches include grid- and genetic-based searches~\cite{HAlibarahim2021}. ML-based methods, such as Bayesian Optimization~\cite{ASouza2020,FHutter2019}, are also applicable. Typically, ML-based HPO methods tune the hyperparameters using the accuracy results obtained from iterative DNN training on the entire dataset. Since it is infeasible to train a model on the entire dataset for each set of global parameters int the resource-constrained edge execution environment, tuning the global parameters round-by-round has been considered as a common practice for FL~\cite{ZMa2021,MKhodak2021}. However, round-by-round global parameter tuning is still challenging due to the following unique aspects of FL:
\vspace{-0.1cm}
\begin{itemize}
    \item {\bf System Heterogeneity and Runtime Variance.} There exist a variety types of system-on-chips (SoCs) with distinct computing performance at the edge~\cite{CJWu2019}, which results in large performance gaps across participating devices. Furthermore, stochastic runtime variance, including on-device interference~\cite{DShingari2015} and network stability~\cite{YGKim2017_2}, can even exacerbate the performance variability~\cite{BGaudette2016,Gaudette:TMC19} across the devices. This results in the straggler problem, where the training time per aggregation round is determined by the slowest device, making it difficult to find the optimal global parameters for each round. 
    \item {\bf Data Heterogeneity.} For model convergence, ensuring training data are independently and identically distributed (IID) for each and every participating device is crucial. However, in edge execution environment, client training data are not guaranteed to be non-IID, as training samples of an individual user are often not representative of the entire population~\cite{ZChai2019,Maeng:RecSys22}. The inclusion of non-IID data in training can defer model convergence~\cite{XLi2020,HBMcMahan2017}. Since global parameters influence the degree of non-IID data reflected in the model gradients; it is also crucial to adjust the parameters considering data heterogeneity.
\end{itemize}

The high degree of heterogeneity and runtime variance makes it challenging to optimize global parameters using offline, server-based simulations. Static simulation-based optimization studies cannot adapt to dynamic system/data heterogeneity or runtime variance. To identify efficient global parameters for each FL training round under system/data heterogeneity, several approaches have been proposed recently~\cite{ZMa2021,MKhodak2021}. However, these prior approaches do not consider the stochastic nature of edge computing, including performance interference and network variability. In addition, prior work did not consider optimizing the energy efficiency of FL, which can lead to increasing energy footprint at scale~\cite{CJWu2022}. To the best of our knowledge, this is the first work to tackle energy-efficient global parameter optimization for FL.

This paper proposes an FL global parameter optimization framework based on reinforcement learning --- FedGPO --- that dynamically adjusts the global parameters (\textit{B, E, K}) to maximize the FL energy efficiency guaranteeing model quality. The optimization is performed round-by-round over the entire training process, considering system and data heterogeneity, as well as runtime variance. Since the optimal (\textit{B, E, K}) varies with the computation characteristics of neural networks (NN), performance profiles of participant device systems, local training sample distributions, and runtime variance, the enormous design space makes it difficult to enumerate exhaustively. Hence, we propose a technique based on a reinforcement learning to addres this optimization formulation. FedGPO identifies the characteristics of NNs and profiles of devices such as the intensity on-device interference, network instability, and the distributions of data samples every aggregation round. It then determines the global parameters for the round, which  maximizes energy efficiency while not deteriorating the training accuracy. Based on the result of the decision, FedGPO continuously learns and predicts the efficient global parameters. The key contributions of this work are as follows:
\vspace{-0.1cm}
\begin{itemize}
\item We present performance and energy efficiency characterization for the FL global parameter design space. The characterization results show that optimal settings vary across the FL training rounds due to the varying level of the system/data heterogeneity and the stochastic runtime variance (Section~\ref{sec:motivation}).
\item We propose a global parameter optimization framework, FedGPO, that identifies the near-optimal global parameter setting for each round, enabling energy-efficient federated learning (Section~\ref{sec:design}).
\item We implement and evaluate FedGPO for FL use cases with 200 mobile devices encompassing three performance categories: high, medium, and low (Section~\ref{sec:result}). Real system-based experiments demonstrate that FedGPO improves the energy efficiency of the participant devices by 3.6x, while satisfying the accuracy requirements.
\end{itemize}
%\vspace{-0.2cm}
\section{Background and Motivation}
\label{sec:motivation}

\subsection{Global Impact of FL Parameters}
\label{sec:motivation1}

\begin{algorithm}[t]
\caption{FedAvg}
\label{alg:FedAvg}
\begin{flushleft}
\textbf{Variable:} \textit{B}, \textit{E}, \textit{K} \\
\hspace*{\algorithmicindent}\hspace*{\algorithmicindent} \textit{B} is the local minibatch size \\
\hspace*{\algorithmicindent}\hspace*{\algorithmicindent} \textit{E} is the number of local epochs \\
\hspace*{\algorithmicindent}\hspace*{\algorithmicindent} \textit{K} is the number of participant devices \\
\textbf{Constants:} \textit{N}, \textit{$\eta$} \\
\hspace*{\algorithmicindent}\hspace*{\algorithmicindent} \textit{N} is the number of entire devices \\
\hspace*{\algorithmicindent}\hspace*{\algorithmicindent} \textit{$\eta$} is the learning rate \\
\textbf{Server executes:} \\
\hspace*{\algorithmicindent}\hspace*{\algorithmicindent} initialize $(B, E, K)$\\
\hspace*{\algorithmicindent}\hspace*{\algorithmicindent} initialize $w_{0}$\\
\hspace*{\algorithmicindent}\hspace*{\algorithmicindent} \textbf{for} each round \textit{t} = 1,2,... \textbf{do}\\
\hspace*{\algorithmicindent}\hspace*{\algorithmicindent}\hspace*{\algorithmicindent} $S_{t}$ $\leftarrow$ (random set of \textit{K} clients among \textit{N} clients) \\
\hspace*{\algorithmicindent}\hspace*{\algorithmicindent}\hspace*{\algorithmicindent} \textbf{for} each client \textit{k} $\in$ $S_{t}$ \textbf{in parallel do} \\
\hspace*{\algorithmicindent}\hspace*{\algorithmicindent}\hspace*{\algorithmicindent}\hspace*{\algorithmicindent} $w^{k}_{t+1}$ $\leftarrow$ ClientUpdate($k$, $w_{t}$) \\
\hspace*{\algorithmicindent}\hspace*{\algorithmicindent}\hspace*{\algorithmicindent} $w_{t+1}$ $\leftarrow$ $\sum_{k=1}^{K} \frac{n_{k}}{n} w_{t+1}^{k}$ \\
\hspace*{\algorithmicindent} \\
\textbf{ClientUpdate(}\textit{k, w}\textbf{):} // \textit{Run on client k} \\
\hspace*{\algorithmicindent}\hspace*{\algorithmicindent} $B$ $\leftarrow$ (split $P_{k}$ into batches of size \textit{B} \\
\hspace*{\algorithmicindent}\hspace*{\algorithmicindent} \textbf{for} each local epoch \textit{i} from 1 to \textit{E} \textbf{do}\\
\hspace*{\algorithmicindent}\hspace*{\algorithmicindent}\hspace*{\algorithmicindent} \textbf{for} batch $b \in B$  \textbf{do}\\
\hspace*{\algorithmicindent}\hspace*{\algorithmicindent}\hspace*{\algorithmicindent}\hspace*{\algorithmicindent} $w$ $\leftarrow$ $w - \eta \nabla \ell (w;b)$ \\
\hspace*{\algorithmicindent}\hspace*{\algorithmicindent} return \textit{w} to server\\
\end{flushleft}
\end{algorithm}

To prevent privacy leaks in ML training, FL is proposed, where edge devices train a shared global model collaboratively without sharing the on-device data samples with the cloud~\cite{KBonawitz2019,JKonecny2016,HBMcMahan2017}. FedAvg is the de-facto FL algorithm~\cite{JKonecny2016,HBMcMahan2017} (Algorithm~\ref{alg:FedAvg}).
For \textit{N} devices, the server initializes a global model along with the number of local training epochs \textit{E}, the local training minibatch size \textit{B}, and the number of participant devices \textit{K}. It also initializes the model parameters $w_{0}$.
In every round \textit{t}, the server randomly selects \textit{K} devices among the \textit{N} devices and transmits the global model to them. Each selected device trains the model locally using the on-device data with a batch size of \textit{B} over \textit{E} epochs. After the local execution of the training is finished, each device transmits the model parameters back to the server. The server then updates the global model with the average of local parameters.  

\begin{figure}[t]
    \centering
    \includegraphics[width=\linewidth]{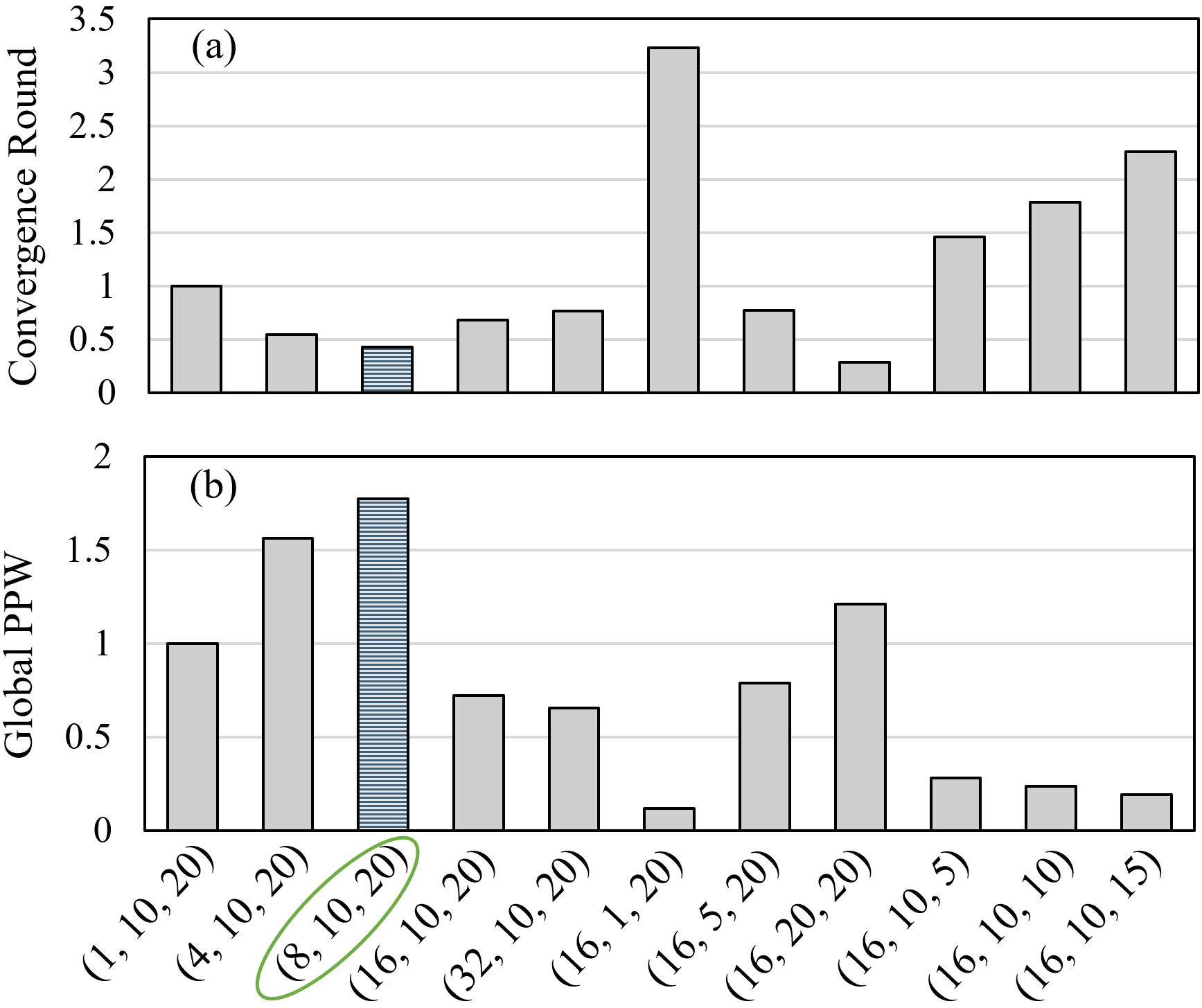}
    \vspace{-0.3cm}
    \caption{Depending on the global parameters, the FL convergence performance and global energy efficiency vary significantly. The convergence round and global PPW are normalized to (1, 10, 20).}
    \label{FL-global-parameter}
    \vspace{-0.2cm}
\end{figure}

Global parameters (\textit{B, E, K}) significantly impact the FL model convergence and energy efficiency. It is therefore crucial to carefully select the parameters for better model quality and the energy efficiency of participating devices.
Figure~\ref{FL-global-parameter} shows the (a) convergence time and (b) global performance per watt (PPW) of a CNN model with the MNIST dataset (CNN-MNIST)~\cite{YLeCun1998,JTSpringenberg2015} for varying FL settings of (\textit{B, E, K}). 

\textit{B} determines the number of local data samples used for one iteration of training on each device. Typically, \textit{B} is associated with the generalization problem --- using larger batch sizes usually yield poor generalizability~\cite{SLSmith2018,EHoffer2017}. For this reason, \textit{B} largely affects the model convergence and global energy efficiency, as shown in Figure~\ref{FL-global-parameter}.

\textit{E} represents the number of training iterations for each device with the same data samples. Since \textit{E} is related to the over- versus under-fitting to specific data samples~\cite{HBMcMahan2017}, it also has a global impact on the model convergence, as shown in Figure~\ref{FL-global-parameter}.

The number of participant devices \textit{K} can be considered as the global batch size in FL, as it is related to the amount of global data used per round. Although smaller values of \textit{K} enable efficient FL deployment by reducing the impact of communication overhead~\cite{HBMcMahan2017}, a careful selection is still required --- \textit{K} is also associated with the generalization problem~\cite{HBMcMahan2017}, affecting the model convergence and global energy efficiency as shown in Figure~\ref{FL-global-parameter}.

\begin{figure}[t]
    \centering
    \includegraphics[width=\linewidth]{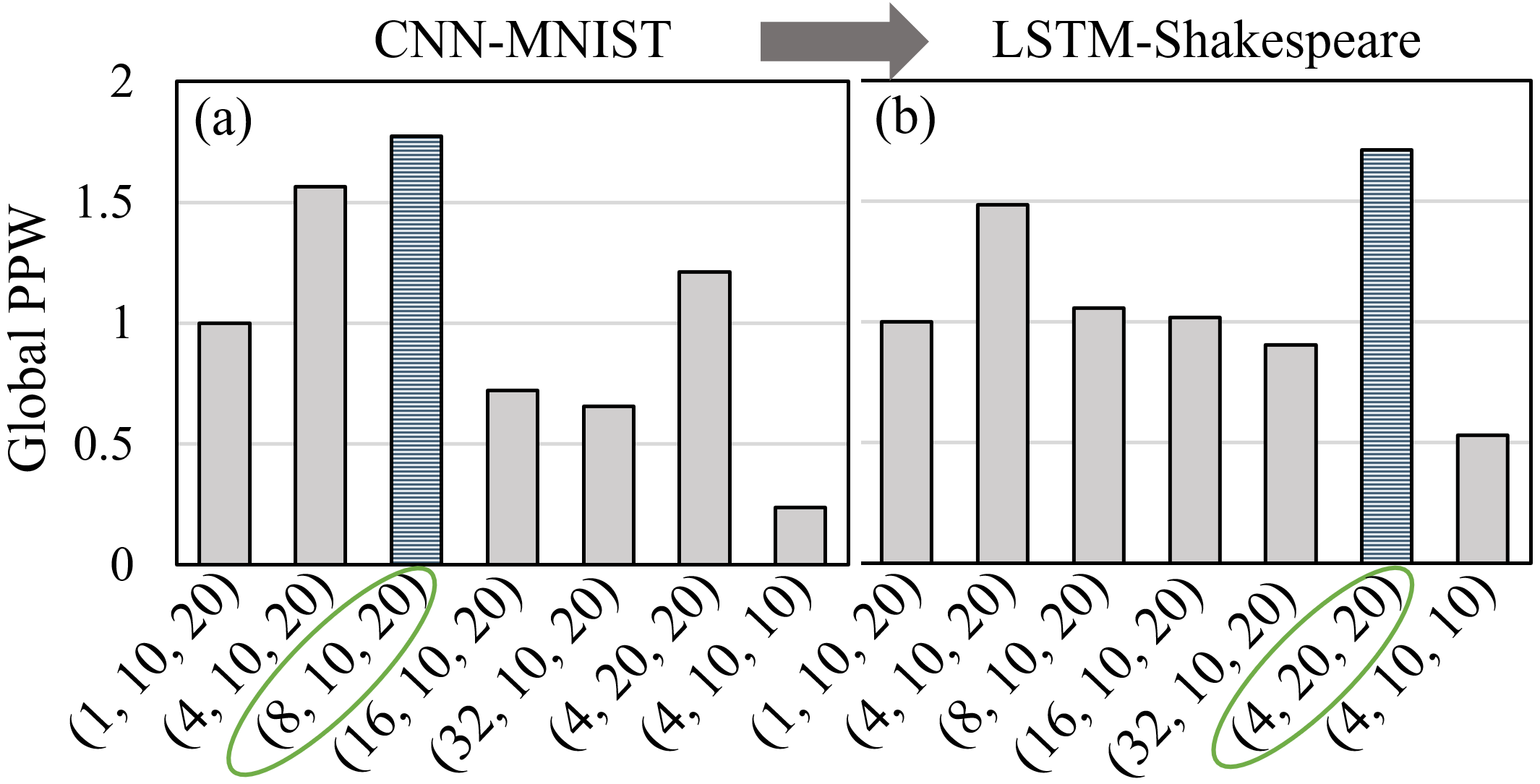}
    \vspace{-0.3cm}
    \caption{The most energy-efficient global parameter combination shifts in accordance with the NN characteristics. The convergence round and global PPW are normalized to (1, 10, 20).}
    \label{FL-NN-characteristics}
    \vspace{-0.4cm}
\end{figure}

It is also important to consider the NN characteristics when selecting the global parameters, as the two are interrelated. Figure~\ref{FL-NN-characteristics} shows the global PPW of two NNs under the different FL settings of (\textit{B, E, K}). In case of CNN-MNIST~\cite{YLeCun1998,JTSpringenberg2015}, (\textit{B, E, K}) of (8, 10, 20) shows the best energy efficiency among the selected global parameter combinations. In contrast, when we use the LSTM model with the Shakespeare dataset (LSTM-Shakespeare)~\cite{HBMcMahan2017,JKonecny2016}, the best energy efficient global parameter combination shifts to (4, 20, 20) as the learning characteristics of LSTM-Shakespeare differ from those of CNN-MNIST. Furthermore, LSTM-Shakespeare comprises more memory-intensive RC layers, whereas CNN-MNIST mainly consists of computation-intensive convolutional and fully-connected layers. Owing to the memory pressure, LSTM-Shakespeare exhibits higher energy efficiency with smaller input batch sizes and more iterations.

\subsection{FL Global Parameter Optimization}
\label{sec:motivation2}

\begin{figure}[t]
    \centering
    \includegraphics[width=\linewidth]{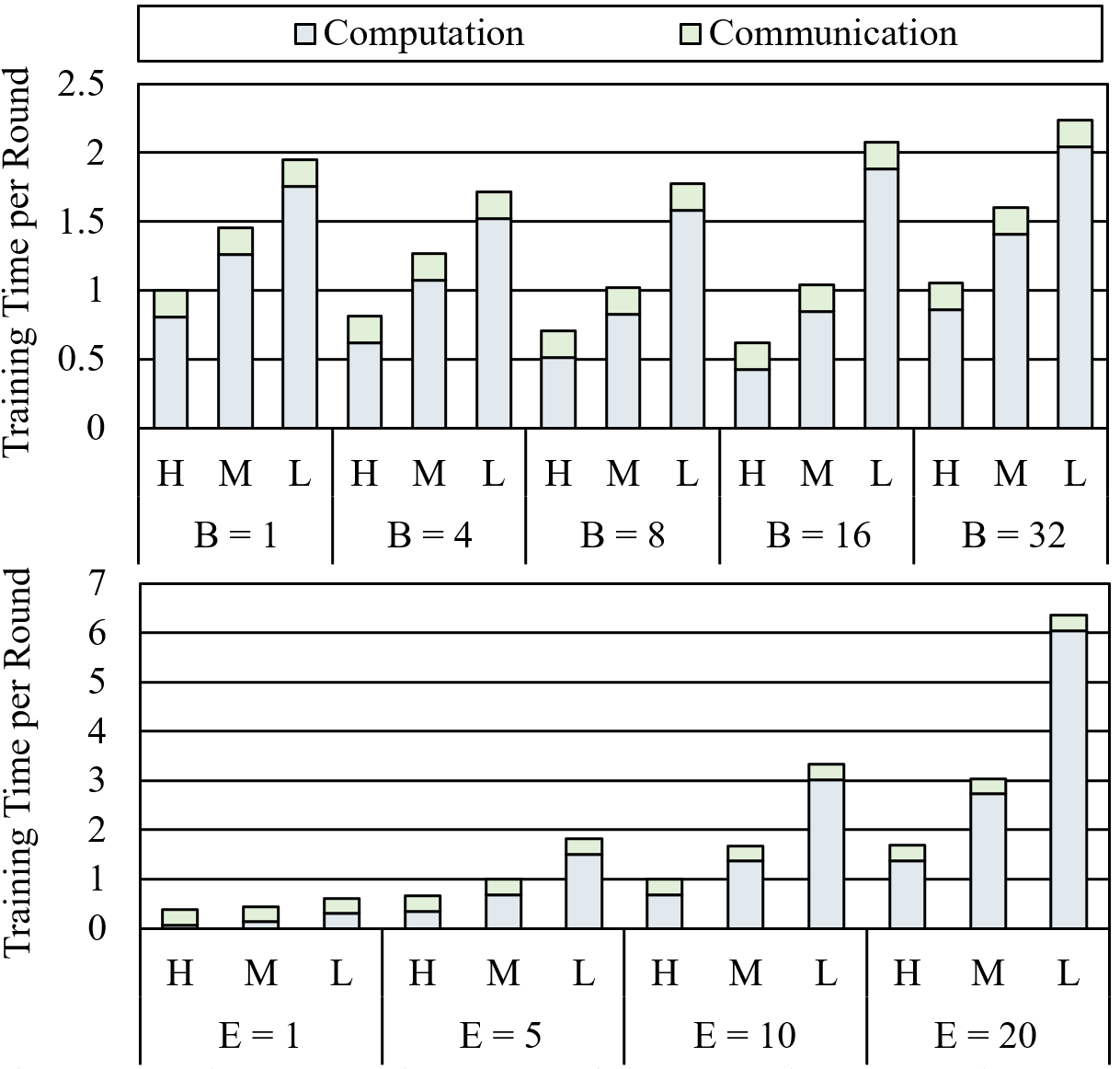}
    \vspace{-0.3cm}
    \caption{Training time per round of devices significantly varies with (a) \textit{B} and (b) \textit{E} introducing large performance gaps across the devices. The training time per round is normalized to that of H with \textit{B} of 1 and that of H with \textit{E} of 10 for (a) and (b), respectively.}
    \label{FL-straggler}
    \vspace{-0.2cm}
\end{figure}

In traditional centralized training, hyperparameter optimization (HPO) is the process of finding a set of hyperparameters that minimizes loss or maximizes accuracy for a DNN~\cite{ASouza2020}. Generally, HPO first randomly selects a set of hyperparameters, and trains the DNN using the entire dataset. The training results (i.e., loss and accuracy) obtained with the selected hyperparameters are measured, and used to select the next set of hyperparameters. Since the hyperparameters can be any integer or floating point values, their search space is usually large. Hence, to efficiently explore the search space, machine learning-based optimization techniques, such as Bayesian Optimization (BO) or Tree Parzan Estimator (TPE), are widely employed] for the HPO process. 

Despite a variety of previous works on centralized training, global parameter optimization in FL has unique challenges:
%. Devices participating in FL are resource constrained, so that evaluating each and every set of global parameters using the entire data set on-device is infeasible. Thus, adjusting the global parameters round-by-round while guaranteeing the convergence has been actively explored for FL recently~\cite{ZMa2021,JZhang2021}. However, tuning the FL global parameters round-by-round is still challenging due to the following unique aspects:

\textbf{Straggler Problem:} For each device, the training time of each device substantially varies with the global parameters, introducing the straggler problem --- the overall training time per round is determined by the slowest device. Figure~\ref{FL-straggler} illustrates the training time per round for different device categories (i.e., H, M, and L for high-end, mid-end, and low-end devices, respectively\footnote{A detailed specification of each device category is presented in Section~\ref{sec:methodology1}.}), depending on different \textit{B} and \textit{E} values. As \textit{B} determines the amount of on-device data to be processed in each iteration, training time on each device category significantly depends on its computation- and memory-capabilities. In addition, as \textit{E} determines the number of iterations, it has a linear impact on the training time per round. 

\begin{figure}[t]
    \centering
    \includegraphics[width=\linewidth]{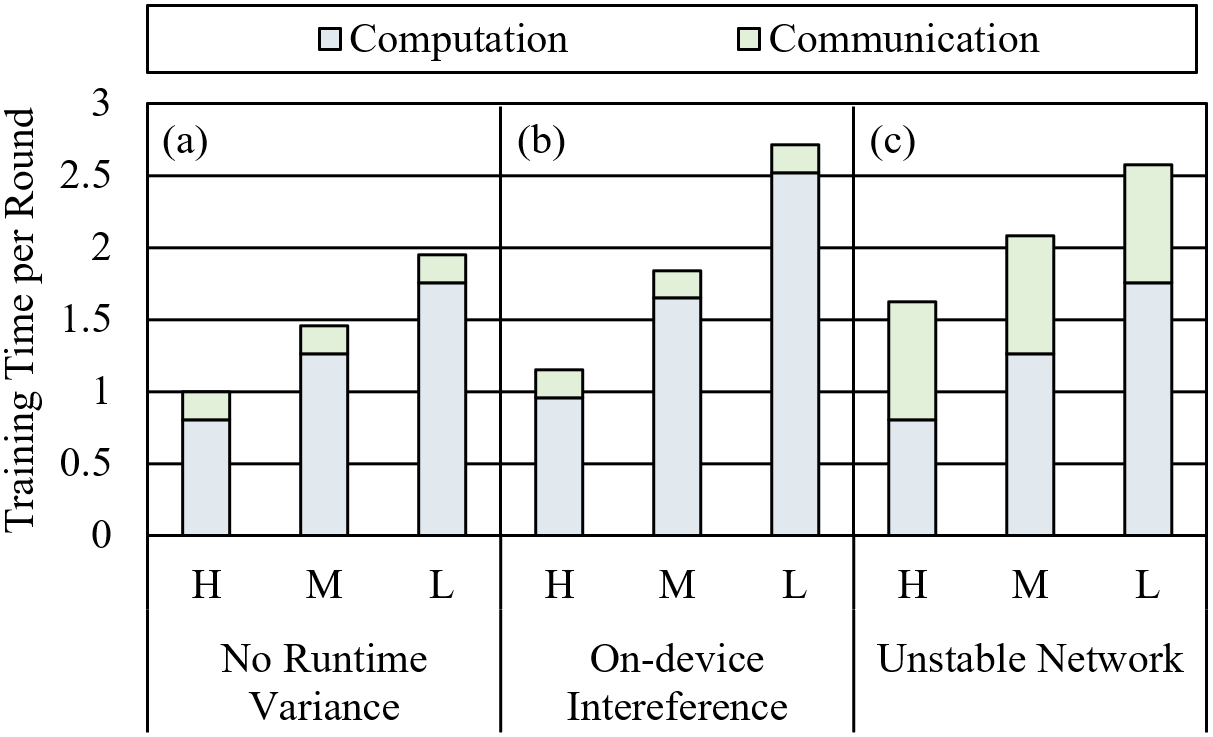}
    \vspace{-0.3cm}
    \caption{Runtime variance significantly affects computation and communication time exacerbating the straggler problem. The training time per round is normalized to that of H in the absence of runtime variance.}
    \label{FL-runtime-variance}
\end{figure}

Stochastic runtime variance exacerbates the straggler problem. In a real use case, there can be several applications co-running with the FL execution, since modern mobile device support multi-tasking features~\cite{DShingari2018}. This causes on-device shared resource interference~\cite{YGKim2017_1,YGKim2020,SYLee2017} degrading computation performance of FL. In addition, signal strength variations in wireless network can affect the performance and energy efficiency of global aggregations in FL --- the data transmission latency and energy increase exponentially at weak signal strength~\cite{YGKim2019,NDing2013}.

Figure~\ref{FL-runtime-variance} shows the training time per round on different device categories, (a) when there is no runtime variance, (b) when there is on-device interference, and (c) when the network is not stable. As shown in Figure~\ref{FL-runtime-variance}(a) and (b), the on-device interference deteriorates the computation time for each device. Since the impact of interference depends on the capabilities of each device in terms of computation and memory, it exacerbates the inter-device performance gaps. Further, as shown in Figure~\ref{FL-runtime-variance}(a) and (c), the network instability deteriorates the communication time of each device, thus affecting the percentage of performance gaps across the devices.

\begin{figure}[t]
    \centering
    \includegraphics[width=\linewidth]{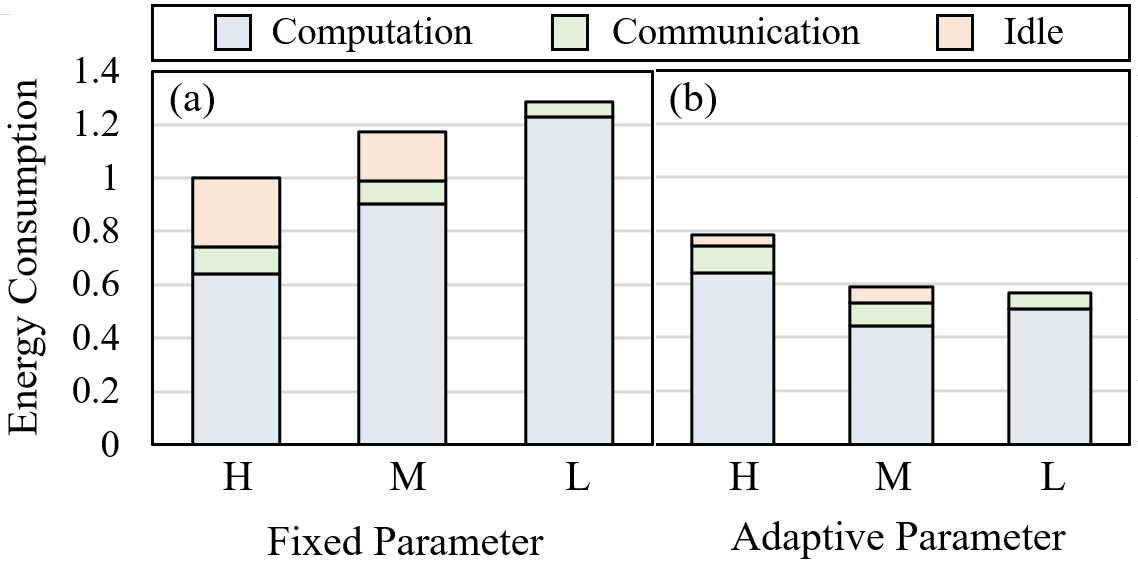}
    \vspace{-0.3cm}
    \caption{The adaptive adjustment of global parameters resolves the straggler problem saving energy consumption of each device category. The energy consumption is normalized to H with fixed parameters.}
    \label{FL-adaptive-parameter}
    \vspace{-0.2cm}
\end{figure}

\begin{figure}[t]
    \centering
    \includegraphics[width=\linewidth]{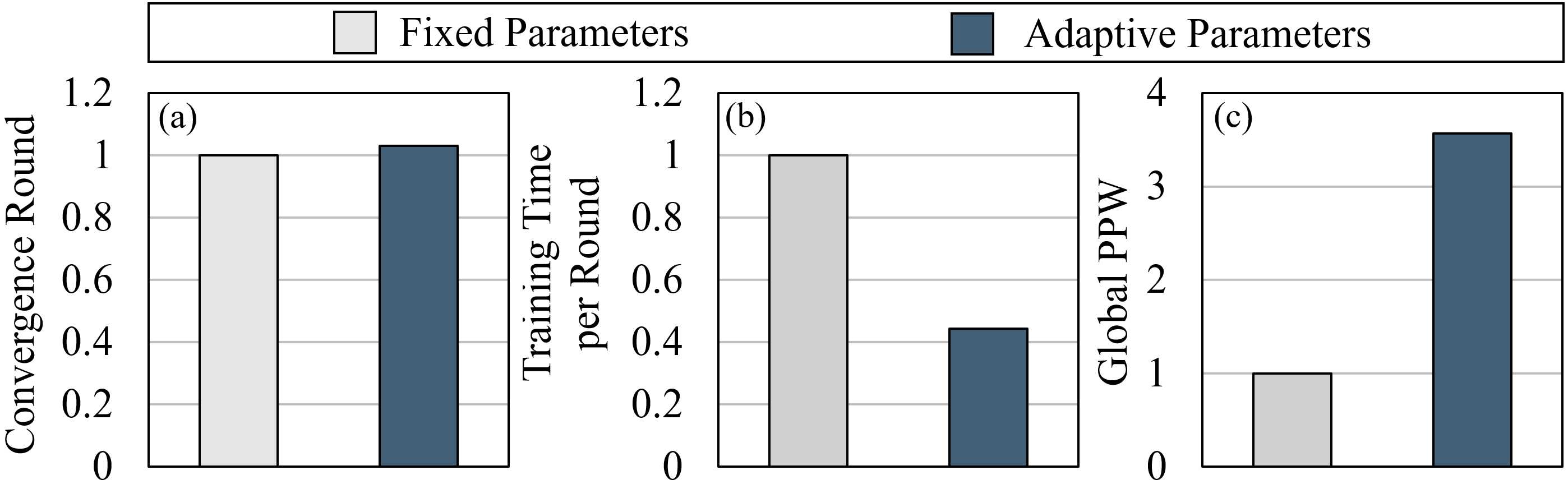}
    \vspace{-0.3cm}
    \caption{Adaptive parameters can improve global PPW by resolving the straggler problem while guaranteeing model convergence.}
    \label{FL-adaptive-parameter2}
    \vspace{-0.2cm}
\end{figure}

The adaptive round-by-round adjustment of the global parameters for different devices can resolve the straggler problem, improving the FL energy efficiency. Figure~\ref{FL-adaptive-parameter} shows the energy consumption of each device (a) when using the same fixed parameters and (b) when adaptively adjusting the parameters for different devices round-by-round. In the former case, faster devices (e.g., H and M) need to wait for the slower devices (e.g., L) consuming energy, as shown in Figure~\ref{FL-adaptive-parameter}(a). Using smaller \textit{B} or \textit{E} for the slower devices can reduce the performance gaps across the devices, saving the energy as shown in Figure~\ref{FL-adaptive-parameter}(b). This significantly improves the average training time per round (2.3x) and global PPW (3.6x), as shown in Figure~\ref{FL-adaptive-parameter2}(b) and (c) respectively --- a careful adjustment of the parameters can still guarantee the model convergence as shown in Figure~\ref{FL-adaptive-parameter2}(a).

\begin{figure}[t]
    \centering
    \includegraphics[width=\linewidth]{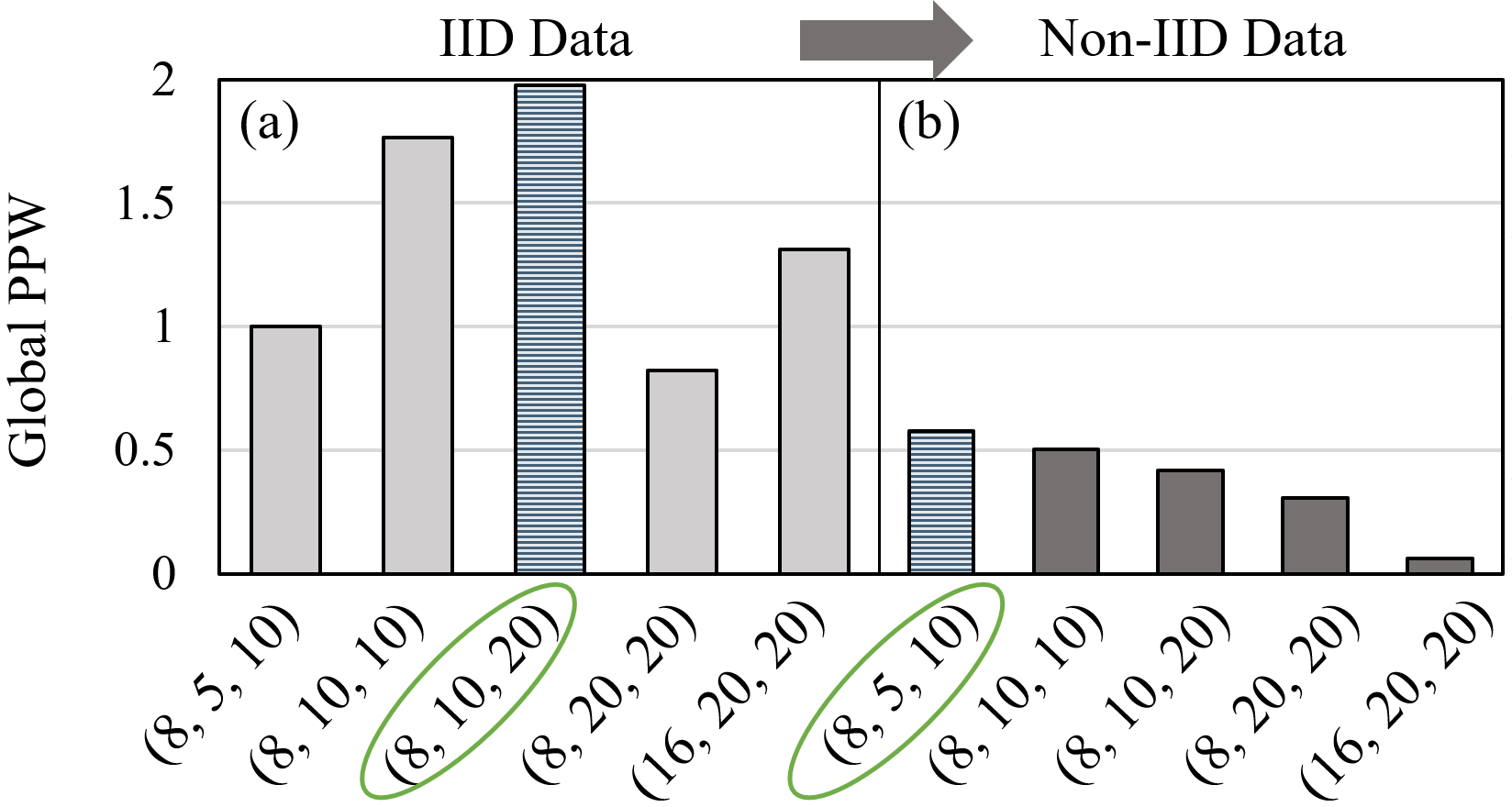}
    \vspace{-0.3cm}
    \caption{The optimal global parameters shift depending on the presence of data heterogeneity.}
    \label{FL-data-heterogeneity}
    \vspace{-0.2cm}
\end{figure}

\textbf{Data Heterogeneity:} It is also crucial for parameter optimization strategies to consider the impact of data heterogeneity. Figure~\ref{FL-data-heterogeneity}(a) shows the global PPW over different global parameter (\textit{B, E, K}) setting, in the absence of data heterogeneity. In this case, the most energy-efficient (\textit{B, E, K}) is (8, 10, 20). In the presence of data heterogeneity, however the global energy efficiency of all (\textit{B, E, K}) is degraded as shown in Figure~\ref{FL-data-heterogeneity}(b), since the data heterogeneity significantly affects model convergence~\cite{YGKim2021}. In this case, the most energy-efficient (\textit{B, E, K}) shifts to (8, 5, 10), as decreasing \textit{E} or \textit{K} reduces the amount of non-IID data reflected to the model parameters --- \textit{E} affects the number of iterations for parameter updates with the given data and \textit{K} affects the number of non-IID devices participating for gradient updates. Since the degree of data heterogeneity can vary round-by-round depending on the participant compositions, it is also important to carefully tune the global parameters round-by-round taking into account the data heterogeneity.
\vspace{-0.1cm}
\section{FedGPO}
\label{sec:design}

\begin{figure}[t]
    \centering
    \includegraphics[width=\linewidth]{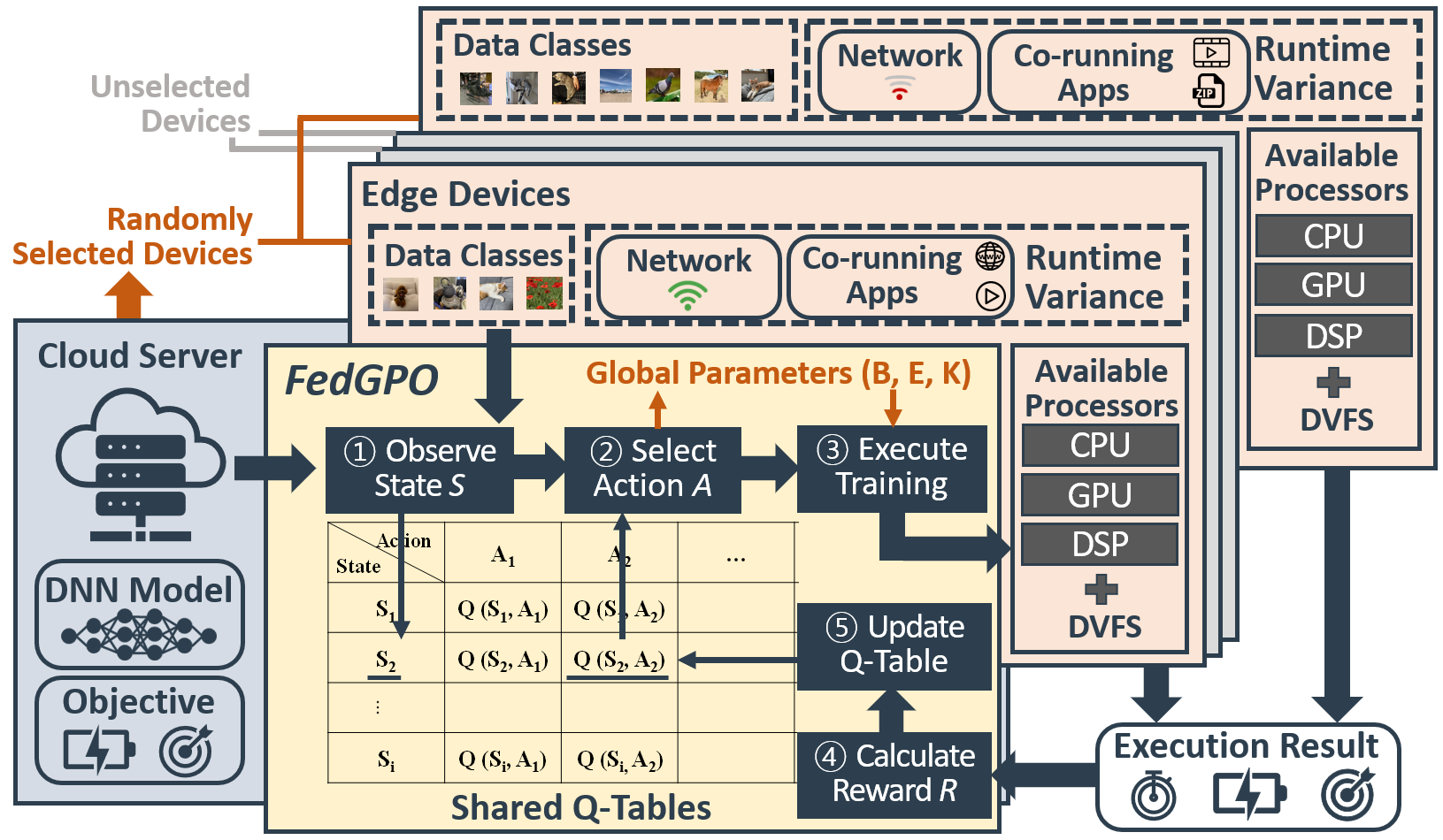}
    \vspace{-0.4cm}
    \caption{FedGPO Design Overview.}
    \label{fig:design}
    \vspace{-0.4cm}
\end{figure}

Under the heterogeneity and runtime variance, it is infeasible to enumerate the large search space associated with FL global parameter optimization. To efficiently explore the optimization space and accurately predict the optimal global parameters, we propose an approach called FedGPO, based on reinforcement learning (RL). Due to its low complexity yet high sample efficiency~\cite{SCKao2020,DYarats2019}, RL has been widely used for the system optimization in the edge domain~\cite{SPagani2020}. In the following sections, we first provide an overview of the FedGPO design, and then elaborate on its RL design and algorithm.

\vspace{-0.1cm}
\subsection{Overview}
\label{sec:design1}

Figure~\ref{fig:design} presents an overview of the FedGPO design. Based on RL, FedGPO attempts to learn an optimal action decision (i.e., global parameter selection) from prior information based on the current state (i.e., heterogeneity and runtime variance) and the given reward (i.e., the amount of global/local energy efficiency and model accuracy improvement of the selected action). 

In each aggregation round, FedGPO identifies the global execution states of FL (\textcircled{\small{1}}), such as characteristics of neural network model architectures and the composition of randomly-selected \textit{K'} participant devices. Note \textit{K'} is \textit{K} determined in the previous aggregation round. FedGPO then identifies the local execution states of the selected devices (\textcircled{\small{1}}), such as the usage of resources, network instability, and the number of data classes each device has. With the identified state information, FedGPO selects the action (\textcircled{\small{2}}), i.e., sets per-device global parameters expected to improve the FL energy efficiency without degrading the model convergence and accuracy. It selects the actions using lookup tables (i.e., Q-tables shared across the devices in the same performance category)~\cite{EE-Dar2006,DEKoulouriotis2008,SPagani2020}, which store the accumulated rewards of previously selected parameter combinations. Using the selected global parameters, FedGPO executes the training on each selected device (\textcircled{\small{3}}). After the aggregation round ends, FedGPO measures its result (i.e., training time, energy consumption, and test accuracy) for calculating the reward (\textcircled{\small{4}}). Finally, FedGPO updates the Q-tables with the calculated reward (\textcircled{\small{5}}). By repeating the aforementioned process (\textcircled{\small{1}}-\textcircled{\small{5}}), FedGPO learns how to select the optimal global parameters.

%\vspace{-0.1cm}
\subsection{FedGPO RL Design}
\label{sec:design2}

\begin{table}[t]
  \caption{Discrete values for FedGPO states.}
  \centering
  \begin{tabular}{|l|l|}
    \hline
    \textbf{State} & \textbf{Discrete Values} \\ \hline
    \multirow{2}{1.4cm}{$S_{CONV}$} & Small ($<$10), medium ($<$20), large ($<$30), \\
    & larger ($>=$40) \\ \hline
    $S_{FC}$ & Small ($<$10), large ($>=$10) \\ \hline
    $S_{RC}$ & Small ($<$5), medium ($<$10), large ($>=$10) \\  \hline
    \multirow{2}{1.4cm}{$S_{Co\_CPU}$} & None (0\%), small ($<$25\%), medium ($<$75\%), \\ 
    & large ($<=$100\%) \\ \hline
    \multirow{2}{1.4cm}{$S_{Co\_MEM}$} & None (0\%), small ($<$25\%), medium ($<$75\%), \\ 
    & large ($<=$100\%)\\ \hline
    $S_{Network}$ & Regular ($>$40Mbps), bad ($<=$40Mbps) \\ \hline
    $S_{Data}$ & Small ($<$25\%), medium ($<$100\%), large (=100\%)\\ \hline
  \end{tabular}
  \vspace{-0.4cm}
  \label{table:States}
\end{table}

To produce accurate predictions of RL, it is crucial to model the core components in a realistic execution environment: 1) state, 2) action, and 3) reward. This section defines the core components for system energy efficiency optimization of FL.

\textbf{State:} We define the FL execution state based on our observations in Section~\ref{sec:motivation}. 

As demonstrated in Section~\ref{sec:motivation1}, the optimal global parameters depend on the NN characteristics. To model the impact of these characteristics, we identify $S_{CONV}$, $S_{FC}$, and $S_{RC}$ which represent the numbers of convolutional, fully-connected, and recurrent layers, respectively --- the three layers are typically highly interrelated with the training/inference efficiency~\cite{YGKim2021}.

In addition, as shown in Section~\ref{sec:motivation2}, the optimal global parameters are also affected by on-device interference. To model this impact, we identify per-device states of $S_{Co\_CPU}$ and $S_{Co\_MEM}$ which represent the CPU utilization and memory usage of co-running applications, respectively. Because the optimal global parameters also depend on network stability, we identify the per-device network stability of $S_{Network}$ with the wireless network (e.g., Wi-Fi and 5G) bandwidth. We also model the data heterogeneity impact by identifying $S_{Data}$ which represents the number of data classes of each device for the aggregation round.

It is difficult to encode continuous values into the RL lookup table. Thus, we convert continuous values of each state into discrete values, by applying a clustering algorithm~\cite{YChoi2019,YGKim2021}. Table~\ref{table:States} summarizes the discretized values. Note FedGPO can support larger search space by further reducing the search space size with different clustering algorithms.

\textbf{Action:} Actions in RL model the customizable control knobs of the system. In the context of FL global parameter optimization, we define the actions as the selection of global parameters for each aggregation round. For the local mini-batch size of \textit{B}, we define the discrete batch size numbers as a feasible range for resource-constrained edge devices~\cite{HBMcMahan2017,YGKim2021}. We also define the discrete numbers of local epochs \textit{E} and participant devices \textit{K} as ranges based on the de-facto FL algorithm~\cite{HBMcMahan2017} to ensure a balanced computation-communication ratio. For example, larger \textit{E} with smaller \textit{K} typically increases the computation-communication ratio. Table~\ref{table:global_parameter} summarizes the discrete values of all global parameters as actions.

\begin{table}[t]
  \caption{Discrete global parameter values for FedGPO actions.}
  \centering
  \begin{tabular}{|c|c|}
    \hline
    \textbf{Parameter} & \textbf{Discrete Values} \\ \hline
    \textit{B} & \{1, 2, 4, 8, 16, 32\} \\ \hline
    \textit{E} & \{1, 5, 10, 15, 20\} \\ \hline
    \textit{K} & \{1, 5, 10, 15, 20\} \\ \hline
  \end{tabular}
    \vspace{-0.5cm}
  \label{table:global_parameter}
\end{table}

\textbf{Reward:} The reward in RL models the optimization objective. To ensure that FedGPO selects global parameters that maximize energy efficiency without degrading model convergence and accuracy, we define the reward \textit{R} as in~(\ref{eq:reward}). Here,  $R_{energy\_local}$ represents the energy consumption of each participant device, whereas $R_{energy\_global}$ denotes that of all devices. $R_{accuracy}$ is the test accuracy of the NN model, while $R_{accuracy\_prev}$ is that of the prior round. Note, since time-to-convergence is not measurable before the convergence, we substitute it with the improvement in accuracy --- a similar practice has been used for the hyperparameter optimization of ML training~\cite{JSnoek2012}.
\begin{equation}
    \label{eq:reward}
    \begin{aligned}
        &if \;\;\; R_{accuracy} \; - \; R_{accuracy\_prev} \;\; <= \;\; 0,\\
        &\qquad R = R_{accuracy} - 100\\
        &else \\
        &\qquad R = - R_{energy\_global} - R_{energy\_local} \\ 
        &\qquad \;\;\;\;\;\;\;\; + \alpha R_{accuracy} + \beta (R_{accuracy} - R_{accuracy\_prev})\\
    \end{aligned}
\end{equation}
 
Among the encoded reward values, we estimate $R_{energy\_local}$ using a commonly-used energy formulation, such as~\cite{RJoseph2001,YGKIM2015,YGKim2017_2}. 

\begin{algorithm}[t]
\caption{Training the Q-learning model}
\label{alg: training}
\begin{flushleft}
\textbf{Variable:} \textit{S}, \textit{A} \\
\hspace*{\algorithmicindent}\hspace*{\algorithmicindent} \textit{S} is the variable for the state \\
\hspace*{\algorithmicindent}\hspace*{\algorithmicindent} \textit{A} is the variable for the action \\
\textbf{Constants:} \textit{$\gamma$, $\mu$, $\epsilon$} \\
\hspace*{\algorithmicindent}\hspace*{\algorithmicindent} \textit{$\gamma$} is the learning rate \\
\hspace*{\algorithmicindent}\hspace*{\algorithmicindent} \textit{$\mu$} is the discount factor \\
\hspace*{\algorithmicindent}\hspace*{\algorithmicindent} \textit{$\epsilon$} is the exploration probability \\
\textbf{Initialize} $Q(S, A)$ as random values \\
\textbf{Repeat} (whenever an aggregation round begins): \\
\hspace*{\algorithmicindent}\hspace*{\algorithmicindent} Observe state and store in \textit{S} \\
\hspace*{\algorithmicindent}\hspace*{\algorithmicindent} \textbf{if} rand() $< \epsilon$ \textbf{then} \\
\hspace*{\algorithmicindent}\hspace*{\algorithmicindent}\hspace*{\algorithmicindent} Choose action \textit{A} randomly \\
\hspace*{\algorithmicindent}\hspace*{\algorithmicindent}\textbf{else} \\
\hspace*{\algorithmicindent}\hspace*{\algorithmicindent}\hspace*{\algorithmicindent} Choose action \textit{A} which maximizes \textit{Q(S,A)} \\
\hspace*{\algorithmicindent}\hspace*{\algorithmicindent} Run training with global parameters defined by \textit{A}\\
\hspace*{\algorithmicindent}\hspace*{\algorithmicindent} (when local training and aggregation terminate) \\
\hspace*{\algorithmicindent}\hspace*{\algorithmicindent} Obtain $R_{energy\_global}$, $R_{energy\_local}$, and $R_{accuracy}$ \\
\hspace*{\algorithmicindent}\hspace*{\algorithmicindent} Calculate reward \textit{R} \\
\hspace*{\algorithmicindent}\hspace*{\algorithmicindent} Observe new state \textit{S'} \\
\hspace*{\algorithmicindent}\hspace*{\algorithmicindent} Choose action \textit{A'} which maximizes \textit{Q(S',A')} \\
\hspace*{\algorithmicindent}\hspace*{\algorithmicindent} \textit{Q(S,A)} $\leftarrow$ \textit{Q(S,A)} +  $\gamma$[\textit{R} + $\mu$\textit{Q(S',A')} - \textit{Q(S,A)}] \\
\hspace*{\algorithmicindent}\hspace*{\algorithmicindent} \textit{S} $\leftarrow$ \textit{S'} 
\end{flushleft}
\end{algorithm}

For each device, we calculate computation energy, $E_{comp}$, using utilization-based CPU and GPU power models~\cite{RJoseph2001,YGKIM2015}, as in~(\ref{eq:comp}). Here, $E^{i}_{CPU\_core}$ and $E_{GPU}$ represent the energy consumed by the \textit{i}th CPU core and GPU, respectively, $E_{PU\_core}$ is the energy consumed by either of the CPU cores or GPU, $t^{f}_{busy}$ is the time spend in the busy state at frequency \textit{f} and $t_{idle}$ is that in the idle state, and $P^{f}_{busy}$ is the power consumed during $t^{f}_{busy}$ at \textit{f} and $P_{idle}$ is that during $t_{idle}$. Note $P^{f}_{busy}$ and $P_{idle}$ for CPU/GPU are obtained by measuring the corresponding processing unit's power at each frequency.
\begin{equation}
    \label{eq:comp}
    \begin{aligned}
        E_{comp} &= \sum_{i}{E_{CPU\_core}^{i}} + E_{GPU}, \\
        E_{PU\_core} &= \sum_{f}(P_{busy}^{f} \times t_{busy}^{f}) + P_{idle} \times t_{idle} \\
    \end{aligned}
\end{equation}
We also calculate communication energy~\cite{YGKim2017_2}, $E_{comm}$, as in~(\ref{eq:communication}), where $t_{TX}$ is the measured transmission latency of the gradient (or parameter) updates, and $P^{S}_{TX}$ is the power consumed by the device during $t_{TX}$ at signal strength \textit{S}. Note $P^{S}_{TX}$ is obtained based on the measured transmission power consumption of devices at each signal strength.
\begin{equation}
    \label{eq:communication}
    \begin{aligned}
        E_{comm} = &P_{TX}^{S} \times t_{TX} 
    \end{aligned}
\end{equation}
We then calculate the idle energy, $E_{idle}$, for the rest of devices (i.e., devices not participating in the round), as in (\ref{eq:idle}), where $t_{round}$ denotes the training time of the round.
\begin{equation}
    \label{eq:idle}
    \begin{aligned}
        E_{idle} = & P_{idle} \times t_{round}
    \end{aligned}
\end{equation}
Based on $E_{comp}$, $E_{comm}$, and $E_{idle}$, $R_{energy\_local}$ is calculated for each device, as in (\ref{eq:Renergy_local}).
\begin{equation}
    \label{eq:Renergy_local}
    \begin{aligned}
        &if \;\; device \;\; \subset \;\; S_{t} \\
        &\qquad R_{energy\_local} = E_{comp} + E_{comm}\\
        &else \\
        &\qquad R_{energy\_local} = E_{idle}
    \end{aligned}
\end{equation}
Based on $R_{energy\_local}$, $R_{energy\_global}$ is also calculated for entire \textit{N} devices (\ref{eq:Renergy_global}).
\begin{equation}
    \label{eq:Renergy_global}
    \begin{aligned}
        R_{energy\_global} = & \sum_{i}^{N}R_{energy\_local}
    \end{aligned}
\end{equation}

\subsection{RL Algorithm}
\label{sec:design2}

To enable real-time decision making for each FL round, we employ Q-learning~\cite{YChoi2019}, as it provides the advantage of low latency overhead by employing lookup tables to determine the best action. For RL, it is also crucial to consider the balance between exploitation and exploration to avoid the local optima~\cite{EE-Dar2006,DEKoulouriotis2008,YGKim2021}. To overcome this issue, we use the epsilon-greedy algorithm~\cite{RNishtala2017,SPagani2020} along with the Q-learning, which chooses an action with the highest reward, or a uniformly-random action based on a pre-specified exploration probability ($\epsilon$). 

In Q-learning, the value function $Q(S, A)$ accepts the state \textit{S} and the action \textit{A} as parameters in the form of a lookup table (Q-table). To permit a large number of participants and address the scalability requirement of FL, FedGPO exploits shared Q-tables\footnote{Shared Q-tables can be a potential source of system usage leakage. To overcome this, FedGPO can exploit per-device Q-tables instead, which imporves the prediction accuracy by 2.7\% degrading 12.2\% of convergence overhead.} for devices within the same performance category. Sharing the learned results across the devices can also expedite the design space exploration process~\cite{YGKim2021}, as each client device experiences different level of heterogeneity and runtime variance.

Algorithm~\ref{alg: training} presents the algorithm for training the shared Q-tables. FedGPO first randomly initializes the Q-tables. During each aggregation round, it observes the execution state \textit{S} as identified in Section~\ref{sec:design1}. It then generates a random value and compares it with $\epsilon$ \footnote{We use 0.1 for $\epsilon$. To determine $\epsilon$, we tested the accuracy and convergence overhead of FedGPO with 0.1, 0.5, and 0.9 of $\epsilon$.}. If the value is lower than $\epsilon$, FedGPO selects the per-device global parameters randomly for exploration. Otherwise, it selects \textit{A} with the largest $Q(S, A)$ using the Q-tables. 

After the local training execution in an FL aggregation round is finished using the selected \textit{A}, FedGPO calculates the reward \textit{R} as described in Section~\ref{sec:design1}. FedGPO then identifies the new execution state \textit{S'}, and selects the corresponding \textit{A'} using $Q(S', A')$ for each device. It then updates the $Q(S, A)$ based on the equation in Algorithm~\ref{alg: training}, where $\gamma$ and $\mu$ are hyperparameters of the learning rate and discount factor, respectively --- $\gamma$ and $\mu$ determined based on the sensitivity analysis.

When the learning phase is completed, i.e., the largest $Q(S, A)$ value is converged for each \textit{S}, FedGPO uses the shared Q-tables to select \textit{A} (i.e., global parameters) for each device to maximize $Q(S, A)$ for \textit{S}.
\section{Experimental Methodology}
\label{sec:methodology}

\subsection{System Infrastructure}
\label{sec:methodology1}

\begin{table}[t]
  \caption{System profiles of Amazon EC2 instances.}
  \centering
  \begin{tabular}{|c|c|c|c|}
    \hline
    \multirow{2}{1.0cm}{\textbf{Category}} & \multirow{2}{1.2cm}{\textbf{Instance}} & \textbf{Performance} & \textbf{RAM} \\ 
    & & \textbf{(GFLOPS)} & \textbf{(GB)} \\ \hline
    H & m4.large & 153.6 & 8 \\ \hline
    M & t3a.medium & 80.0 & 4 \\ \hline
    L & t2.small & 52.8 & 2 \\ \hline
  \end{tabular}
  \vspace{-0.4cm}
  \label{table:device}
\end{table}

We emulate FL with 200 mobile devices referring to prior studies pertaining to FL~\cite{TLi2020,HBMcMahan2017,YGKim2021,JKonecny2016}. Since it is difficult to run experiments with 200 real smartphones, we emulate the FL performance by using Amazon EC2 instances~\cite{Amazon2} (Table~\ref{table:device}), which feature equivalent theoretical GFLOP performance and memory capacity to those of the three smartphone performance categories: high-end (H), mid-end (M), and low-end (L) devices. By referring to in-the-field system performance distribution~\cite{CJWu2019}, we composed 200 instances with 30 H, 70 M, and 100 L devices.
For the model aggregation server, we use a c5d.24xlarge Amazon EC2 instance, whose theoretical performance and RAM are 448 GFLOPS and 32GB, respectively. 

For measuring the power, we use three representative smartphones for each performance category~\cite{YGKim2020}: Mi8Pro~\cite{Huawei}, Galaxy S10e~\cite{Samsung}, and Moto X Force~\cite{Motorola} (Table~\ref{table:Devices}). We use an external Monsoon Power Meter~\cite{Monsoon} to measure three smartphones' power consumption while running on-device training (implemented with DL4j~\cite{Dl4j}) --- a similar approach was used in prior studies~\cite{DPandiyan2013,DShingari2015,YGKim2021}. Based on the measured performance and power consumption, we evaluate the energy efficiency of participant devices in FL.

%To characterize the FL energy efficiency with various clusters of participant devices, 
%To characterize the FL energy efficiency, we compare the energy efficiency of the various global parameter combinations in Section~\ref{sec:motivation}. Based on the characterization results, we build FedGPO as described in Section~\ref{sec:design}, and implement it upon the FedAvg algorithm~\cite{JKonecny2016,HBMcMahan2017} using PyTorch~\cite{PyTorch}. 
To evaluate the effectiveness of FedGPO, we implement it on top of the FedAvg algorithm~\cite{JKonecny2016,HBMcMahan2017} with PyTorch~\cite{PyTorch}. We compare FedGPO with three baselines:
\begin{itemize}
    \item Fixed (Best), which uses the most energy-efficient parameter combination identified by grid search, yet fixed during the entire FL rounds.
    \item Adaptive (BO) which adjusts the global parameters each round using a Bayesian Optimization algorithm where many state-of-the-art approaches are based~\cite{ASouza2020}. 
    \item Adaptive (GA) which adjusts the global parameters every round using a genetic algorithm~\cite{HAlibarahim2021}.
\end{itemize} 
We also compare FedGPO with two previous approaches: FedEX~\cite{MKhodak2021} and ABS~\cite{ZMa2021}. FedEX adjusts the parameters with an exponentiated gradient updates, whereas ABS only adjusts the local minibatch size with a Deep RL.

\begin{table}[t]
  \caption{Specifications of mobile devices.}
  \centering
  \begin{tabular}{|c|c|c|}
    \hline
    \textbf{Device}& \textbf{CPU}& \textbf{GPU} \\
    \hline
    \multirow{3}{1.25cm}{Mi8Pro (H)} & Cortex-A75 (2.8GHz) & Adreno 630 (0.7GHz) \\ 
    & 23 V/F steps & 7 V/F steps \\
    & 5.5 W & 2.8 W \\
%    & Cortex A55 -& 7 V/F steps &\\
%    & 1.8GHz -& (2.8 W) &(1.8 W)\\
%    & 18 V/F steps & & \\
%    & (5.5 W) & &\\
    \hline
    \multirow{3}{1.25cm}{Galaxy S10e (M)} & Mongoose (2.7GHz) & Mali-G76 (0.7GHz) \\ 
    & 21 V/F steps & 9 V/F steps \\
    & 5.6 W & 2.4 W \\
    \hline
    \multirow{3}{1.25cm}{Moto X Force (L)} & Cortex-A57 (1.9GHz) & Adreno 430 (0.6GHz) \\ 
    & 15 V/F steps & 6 V/F steps \\
    & 3.6 W & 2.0 W \\
    \hline
  \end{tabular}
    \vspace{-0.2cm}
  \label{table:Devices}
\end{table}

We determine two hyperparameters (i.e., learning rate and discount factor) of FedGPO by evaluating the three values of 0.1, 0.5, and 0.9 for each one~\cite{YChoi2019,YGKim2021}. We find that a higher learning rate improves prediction accuracy, as FedGPO works better when a higher amount of the reward is reflected in the Q-tables --- FedGPO needs to adapt to the heterogeneity and stochastic variance within the limited rounds.
In contrast, we observe that a lower discount factor improves prediction accuracy, as FedGPO exhibits higher performance when a lower amount of the reward or the following state is reflected in that of the current state --- sequential states have a weak mutual relationship because of their stochastic nature. Accordingly, in our evaluation, we use 0.9 and 0.1 for the respective hyperparameter.

\subsection{Workloads and Execution Scenarios}
\label{sec:methodology2}

\textbf{Workloads:} We evaluate FedGPO with two workloads~\cite{TLi2020,HBMcMahan2017,JKonecny2016}: (1) training a CNN model with the MNIST dataset ({\bf CNN-MNIST}) for image classification~\cite{YLeCun1998,JTSpringenberg2015} and (2) training an LSTM model with the Shakespeare dataset ({\bf LSTM-Shakespeare}) for the next character prediction~\cite{HBMcMahan2017,JKonecny2016}. 
We also use a state-of-the-art NN workload: (3) training the MobileNet with the ImageNet dataset ({\bf MobileNet-ImageNet}) for image classification~\cite{JDeng09,AHoward2017}. Note our study focuses on mobile-centric neural networks by referring to recent FL deployment and use cases~\cite{Google_new,Google_new2,Apple-new,Meta,Ahard2018,JZhou2021,Dzmitry:mlsys-papaya}, as larger networks have been infeasible for resource-constrained mobile devices~\cite{FEL:arxiv}.

\textbf{Runtime variance:} To emulate realistic on-device interference, we run a synthetic co-running application on a random subset of devices. The synthetic application exhibits the same CPU and memory usage as the real-world mobile application of web browsing~\cite{DShingari2015,DPandiyan2013}. In addition, as the variability of real-world network follows a Gaussian distribution~\cite{NDing2013,YGKim2017_2}, we generate a random network bandwidth following such a distribution using a Wi-Fi AP.

{\bf Data distribution:} We evaluate different degrees of data heterogeneity with two different data sample distributions~\cite{ZChai2019,XLi2020}: Ideal IID and Non-IID. For Ideal IID, all the data classes are evenly distributed to the devices.
In contrast, for non-IID, each data class is randomly distributed following a Dirichlet distribution with a 0.1 concentration parameter~\cite{YGKim2021,ZChai2019,QLi2021,XLi2020,TLin2020,GLee2021,XMu2021,XWu2022}. 
\section{Evaluation Results and Analysis}
\label{sec:result}

\subsection{Result Overview}
\label{sec:result1}

\begin{figure}[t]
    \centering
    \includegraphics[width=\linewidth]{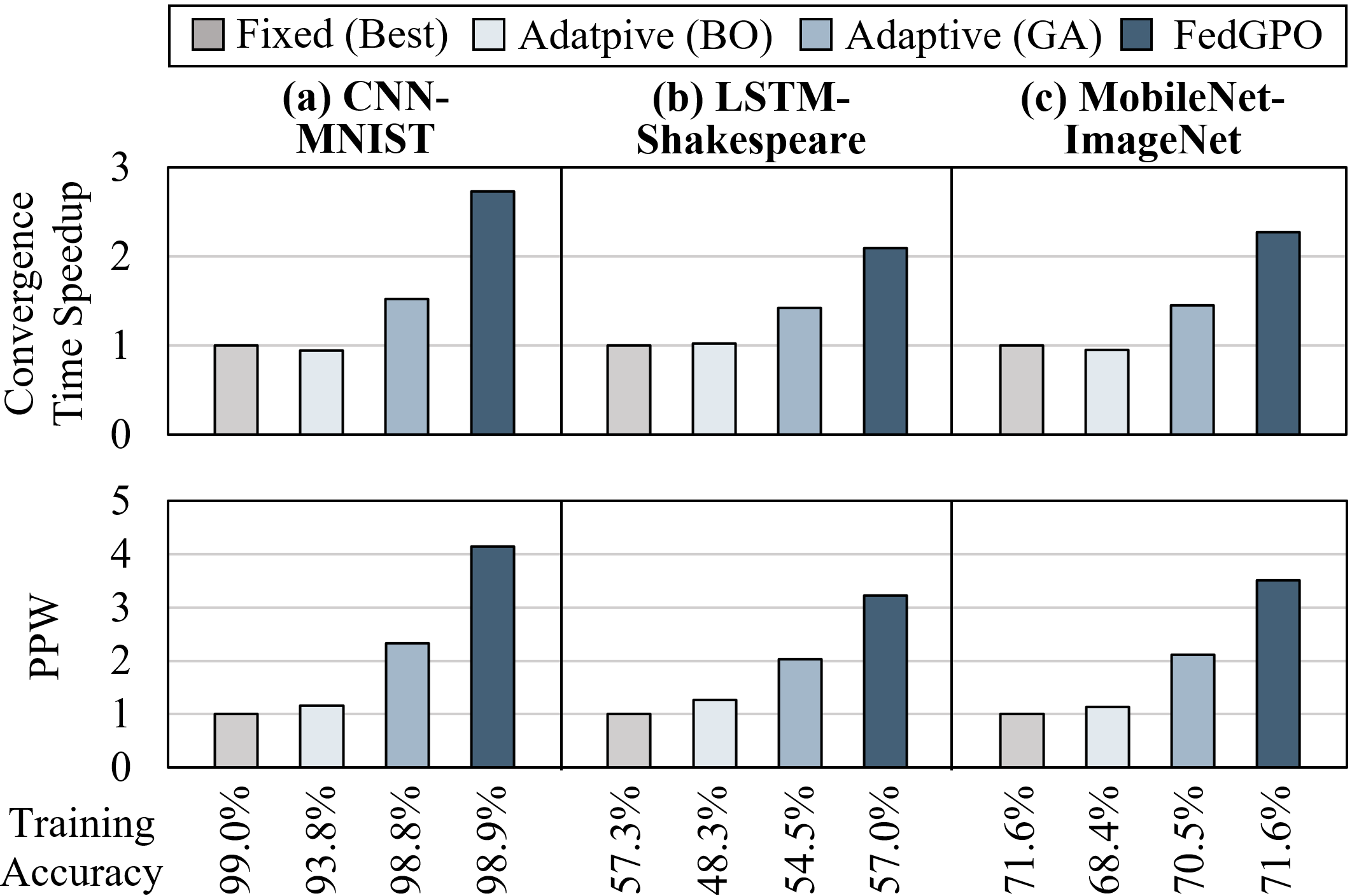}
    \vspace{-0.3cm}
    \caption{FedGPO improves the PPW by 4.1x, 3.2x, and 3.5x compared to the baseline Fixed (Best) for CNN-MNIST, LSTM-Shakespeare, and MobileNet-ImageNet, respectively. It also maintains the training accuracy with the improved convergence time.
    }
    \vspace{-0.4cm}
    \label{fig:overall}
\end{figure}

Figure~\ref{fig:overall} compares the PPW energy efficient, the convergence performance, and the training accuracy among the three FL applications. The PPW and the convergence time speedup are normalized to the Fixed (Best) case. Compared to the Fixed (Best), Adaptive (BO), and Adaptive (GA), FedGPO achieves 3.6x, 3.1x, and 1.7x of the average FL energy efficiency improvement, respectively. It also maintains the training accuracy with the improved convergence time. 

For all FL use cases, FedGPO alleviates the performance gaps across the participant devices, by identifying more efficient per-device global parameters compared to the baseline settings. This improves the average training time per round by 2.3x over the Fixed (Best). Additionally, by reducing the performance gaps across the devices, the redundant energy consumption of the devices is saved by 57.5\% over the Fixed (Best). As a result, significantly better energy efficiency is achieved compared to the baseline settings. 

The global parameters selected by FedGPO also guarantees the convergence (i.e., the training loss settles to a certain value~\cite{TMitchell1997} while the training accuracy gets to an error range of the value achieved by the baseline in an ideal environment). FedGPO maintains the similar number of convergence rounds with Fixed (Best), improving the convergence time while maintaining the model quality; the convergence round difference of Fixed (Best) and FedGPO is only 0.2\%. 

Compared to Adaptive (BO), FedGPO shows 2.4x better convergence time. This is mainly because the Adaptive (BO) has lower sample efficiency than FedGPO, and thus fails to adapt to the heterogeneity and runtime variance round-by-round. GA has a relatively higher sample efficiency than BO~\cite{SCKao2020}, but it still requires a number of mutation/crossover generations for the convergence. Due to the quick adaptability to the hetergeneity and runtime variance every round, FedGPO exhibits 1.6x better convergence time, compared to Adaptive (GA).

\subsection{Adaptability and Accuracy Analysis}
\label{sec:result2}

\begin{figure}[t]
    \centering
    \includegraphics[width=\linewidth]{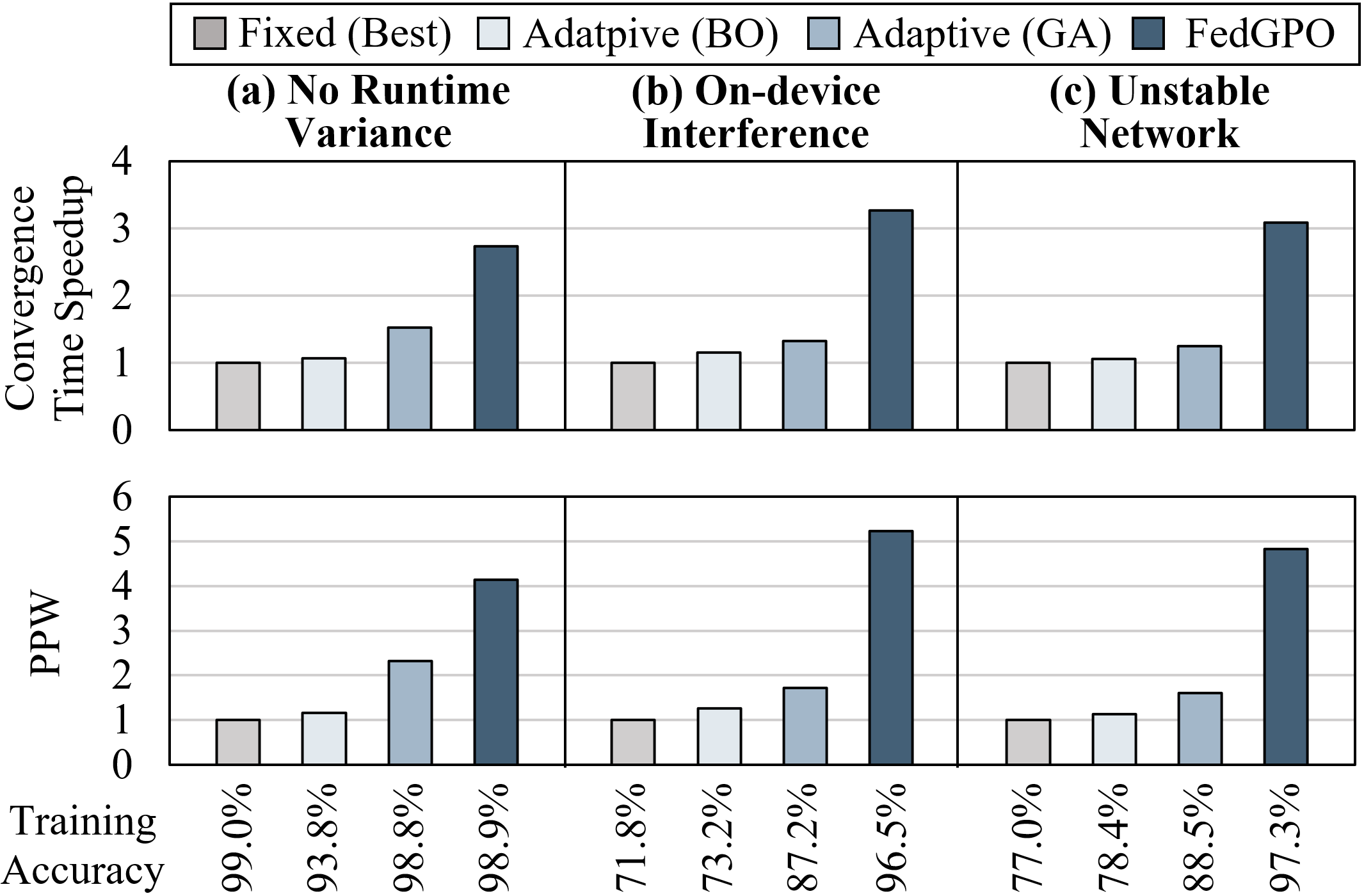}
    \vspace{-0.3cm}
    \caption{
   Under the runtime variance, FedGPO significantly improves the FL energy efficiency by 5.0x, 4.2x, and 3.0x, on average, compared to Fixed (Best), Adaptive (BO), and Adaptive (GA), respectively. It also improves the convergence time maintaining the training accuracy.}
    % Since AutoFL accurately predicts the optimal participants and execution targets in the presence of runtime variance, it improves the energy efficiency and convergence time of FL in realistic environments.}
    \label{fig:realistic environment}
    \vspace{-0.4cm}
\end{figure}

{\bf Adaptability to stochastic variance:} Figure~\ref{fig:realistic environment} compares the PPW, convergence performance, and model accuracy of CNN-MNIST, (a) in the absence of runtime variance, (b) in the presence of on-device interference from co-running applications, and (c) in the presence of network variance. When there exists runtime variance, FedGPO achieves 5.0x, 4.2x, and 3.0x of the average energy efficiency improvement, compared to Fixed (Best), Adaptive (BO), and Adaptive (GA), respectively. Furthermore, it also improves the convergence time while maintaining the training accuracy. Note other NNs show similar result trends.

Under the runtime variance, the performance gap across participating devices significantly varies round-by-round due to the varying on-device computation and communication time. Nevertheless, FedGPO selects better per-device global parameters every round compared to the baseline settings, by quickly adapting varying on-device computation or communication time with high sample efficiency. As a result, the convergence time is improved by 3.2x, 2.9x, and 2.5x, compared with that of Fixed (Best), Adaptive (BO), and Adaptive (GA), respectively. Additionally, by eliminating the redundant energy consumption of the participating devices, FedGPO also significantly improves the energy efficiency compared to the baseline settings. Note, in this case, the training accuracy of the baseline setting is significantly degraded due to the exacerbated straggler problems --- previous works just drop the gradient updates from the stragglers~\cite{QLi2021,TLi2020}.

\begin{figure}[t]
    \centering
    \includegraphics[width=0.9\linewidth]{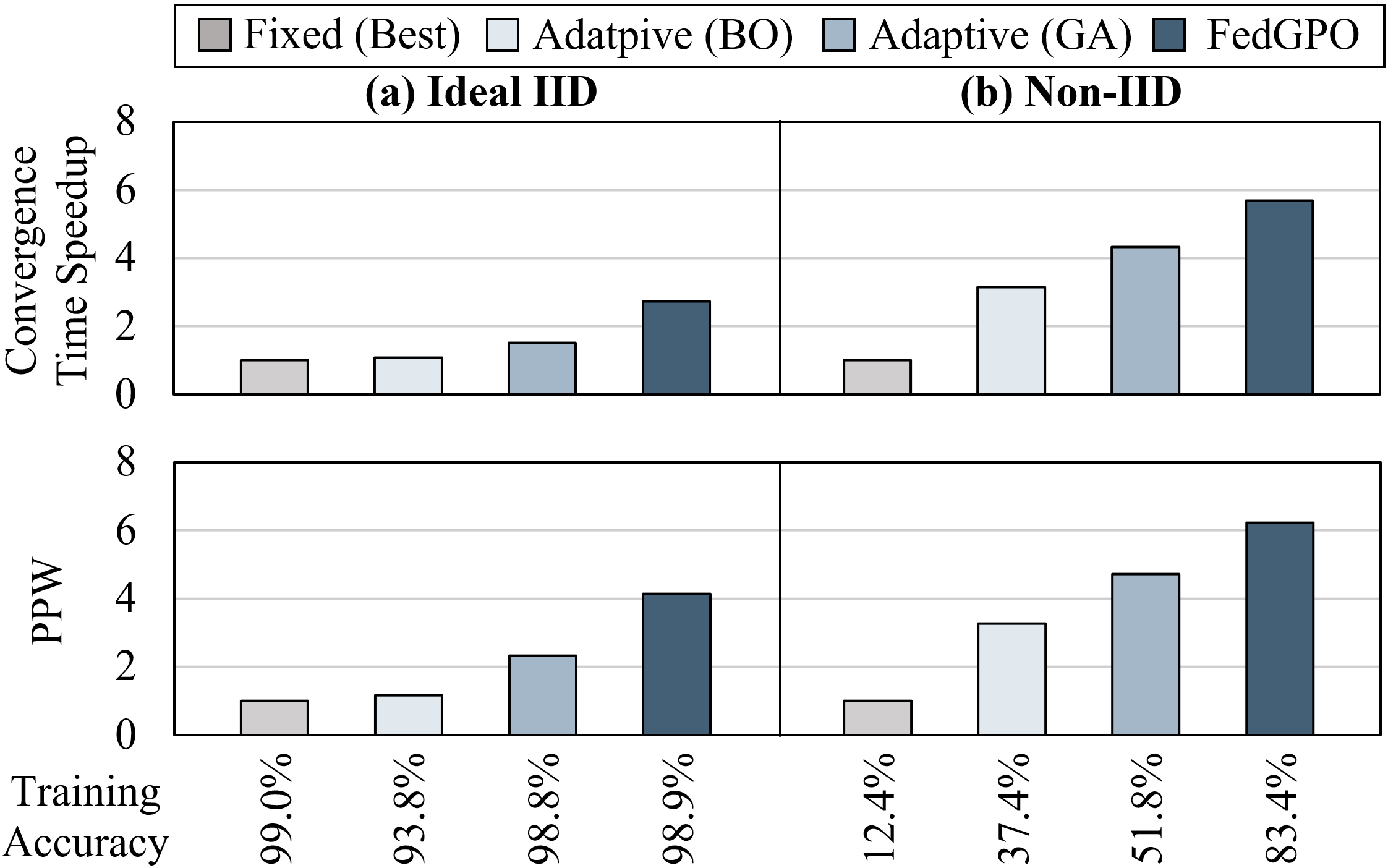}
    \caption{Even in the presence of data heterogeneity, FedGPO  achieves 6.2x, 1.9x, and 1.3x higher energy efficiency than Fixed (Best), Adaptive (BO), and Adaptive (GA), respectively, by adaptively adjusting \textit{E} and \textit{K} along with \textit{B} round-by-round.}
    % Power, and Performance until the last (i.e., 500) round. 
    % In this case, the participating devices keep consuming the energy until the last round.
    %Since AutoFL accurately predicts the optimal participants in the presence of data heterogeneity, it largely improves the energy efficiency and convergence time, showing near-optimal training accuracy. Note missing bars in the upper figure indicate the training model is not converged.
    \label{fig:data}
    \vspace{-0.4cm}
\end{figure}

{\bf Adaptability to data heterogeneity:} Figure~\ref{fig:data} illustrates the energy efficiency, convergence time, and training accuracy of CNN-MNIST, (a) in the absence of data heterogeneity (i.e., Ideal IID) and (b) in the presence of data heterogeneity (Non-IID). Even under the latter scenario, FedGPO still improves the PPW by 6.2x, 1.9x, and 1.3x, compared with Fixed (Best), Adaptive (BO), and Adaptive (GA), respectively. It also improves convergence time and training accuracy against the baseline settings. 

In the presence of non-IID participants, neither \textit{E} nor \textit{K} is adjusted by the baseline settings depending on the degree of data heterogeneity, maintaining the amount of non-IID data reflected on the model gradients. On the other hand, FedGPO learns how data heterogeneity affects the energy efficiency and convergence performance, and adjusts gradient updates with \textit{E} and \textit{K} along with \textit{B}. Therefore, it significantly improves the PPW, convergence performance, and model accuracy even under the data heterogeneity. The prediction accuracy of the baseline settings is also significantly degraded in this case, as they accept the gradient (or parameter) updates from non-IID participants as equally as those from IID-participants~\cite{CWang2020}.

{\bf Prediction accuracy:} FedGPO accurately selects the near-optimal global parameters round-by-round. Table~\ref{table:accuracy} lists the mean absolute percentage accuracy of FedGPO over the optimal global parameters for each round --- these parameters are identified in terms of minimizing the performance gap across the devices, rather than global convergence. FedGPO achieves an average prediction accuracy of 94.7\%.

FedGPO also adapts to the stochastic features of edge execution environments. As shown in Table~\ref{table:accuracy}, in the presence of runtime variance (i.e., on-device interference and network variability), FedGPO successfully selects the near-optimal global parameters, achieving a 94.4\% average prediction accuracy. In the presence of data heterogeneity, FedGPO exhibits relatively lower prediction accuracy (88.9\% on average), as the minimization of the performance gaps across devices does not guarantee the model convergence in this case, whereas FedGPO selects the parameters that guarantee the model convergence while improving the energy efficiency, as shown in Figure~\ref{fig:data}.

\subsection{Comparison with Prior Work}
\label{sec:result3}

\begin{comment}
\begin{figure}[t]
    \centering
    \includegraphics[width=0.95\linewidth]{Figures/Fig.11.png}
    %\vspace{-0.3cm}
    \caption{Compared with FedEX~\cite{MKhodak2021} and ABS~\cite{ZMa2021,JZhang2021}, FedGPO achieves better convergence time and higher energy efficiency.}
    %\vspace{-0.3cm}
    \label{fig:prior}
\end{figure}
\end{comment}

\begin{figure}[t]
    \centering
    \includegraphics[width=0.95\linewidth]{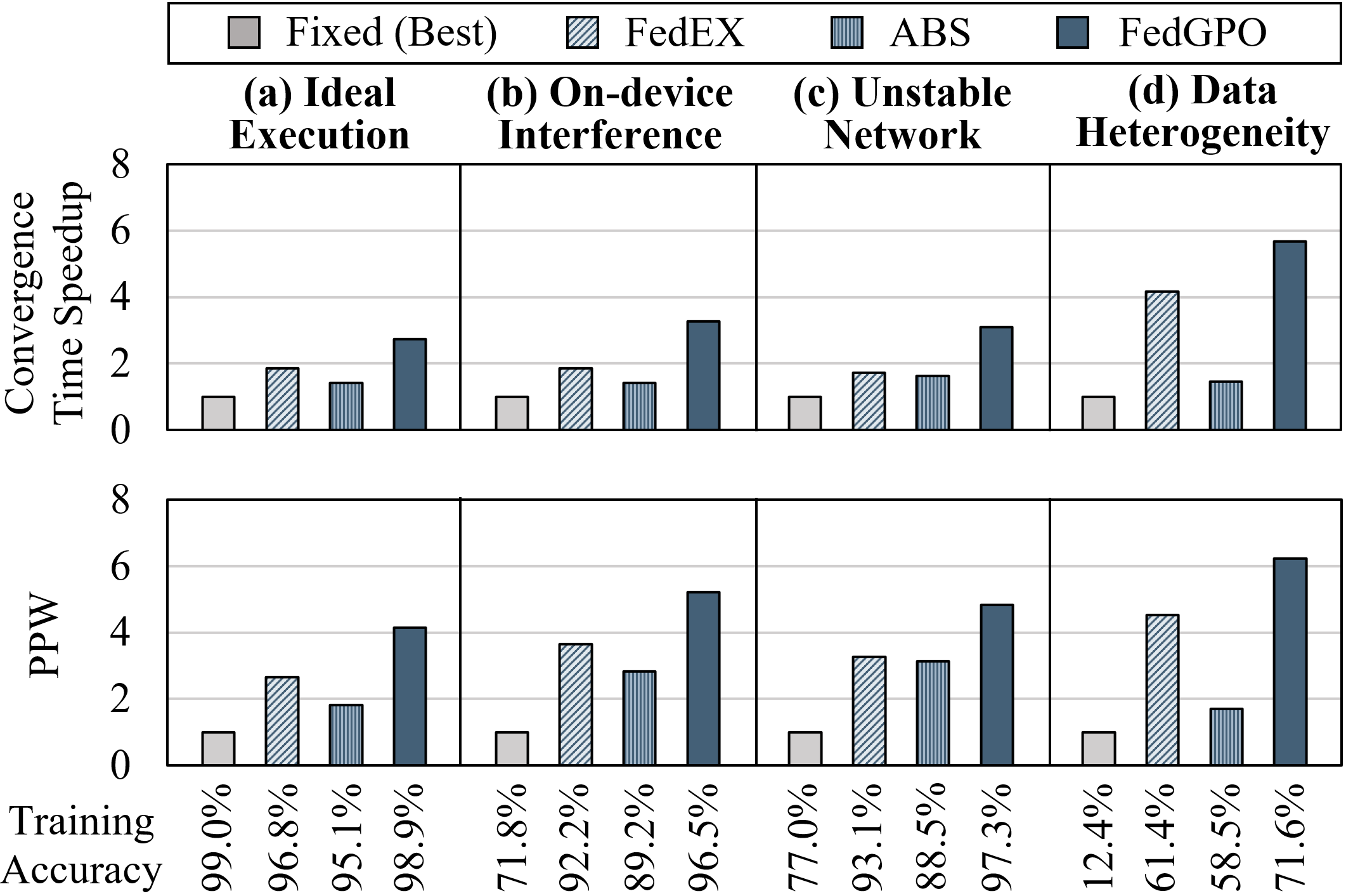}
    %\vspace{-0.3cm}
    \caption{FedGPO outperforms both FedEX~\cite{MKhodak2021} and ABS~\cite{ZMa2021}, in terms of convergence time and energy efficiency, regardless of the presence of runtime variance or data heterogeneity.}
    \vspace{-0.4cm}
    \label{fig:prior2}
\end{figure}

We evaluate FedGPO compared to two prior works for CNN-MNIST: FedEX~\cite{MKhodak2021} and ABS~\cite{ZMa2021}. On average, FedGPO achieves 1.5x and 2.1x improvements in energy efficiency compared to FedEX and ABS, respectively (Figure~\ref{fig:prior2}). Note other NNs show similar result trends.

Under the runtime variance, FedEX and ABS improves the convergence performance and PPW over the baseline, as they reduce the device performance gap, by explicitly considering the straggler problem. However, FedEX does not adapt to the runtime variance as quickly as FedGPO does, due to the lower sample efficiency of exponentiated gradient updates. Furthermore, ABS does not adjust \textit{E} and \textit{K}, which helps to deal with the straggler problem and data heterogeneity. As a result, FedGPO further improves PPW by 1.5x and 1.7x over FedEX and ABS, respectively (Figure~\ref{fig:prior2}).

Compared to the baseline, FedEX is robust to data heterogeneity by adjusting the amount of non-IID data reflected on the model gradients with \textit{E} and \textit{K}. However, it still fails to quickly adapt to the data heterogeneity due to its low sample efficiency. In contrast, ABS is not robust to data heterogeneity, as \textit{B} is not closely related to the data heterogeneity as explained in Section~\ref{sec:motivation}. Consequently, FedGPO achieves 1.4x and 3.6x higher PPW over FedEX and ABS, respectively (Figure~\ref{fig:prior2}).

\subsection{Convergence and Overhead Analysis}
\label{sec:result3}

\begin{scriptsize}
\begin{table}[t]
  \caption{Accuracy for Global Parameter Selection.}
  \centering
  \begin{tabular}{|c|c|c|}
    \hline
    \textbf{Runtime} & \textbf{Data} & \textbf{Prediction} \\
    \textbf{Variance}& \textbf{Heterogeneity} & \textbf{Accuracy} \\ \hline
    No & No & 94.7\% \\ \hline
    Yes (On-device Interference) & No & 94.2\% \\ \hline
    Yes (Unstable Network) & No & 94.5\% \\ \hline
    No & Yes & 87.7\% \\ \hline
    Yes & Yes & 90.1\% \\ \hline
  \end{tabular}
  \vspace{-0.4cm}
  \label{table:accuracy}
\end{table}
\end{scriptsize} 

When training the shared Q-tables of FedGPO, the reward converges after 30-40 aggregation rounds. Prior to convergence, FedGPO shows 24.2\% lower energy efficiency than Fixed (Best), on average. After the convergence, however, FedGPO selects more efficient global parameters than the baselines, as we observed in Section~\ref{sec:result2}. Consequently, the global energy efficiency is eventually improved compared with that of the baselines.

The average runtime cost associated with training the shared Q-tables is 499.6 $\mu$s except for the FL execution time, corresponding to 0.7\% of the training time per round. The overhead includes 1) identifying the per-device states (496.8 $\mu$s), 2) choosing global parameters (0.2 $\mu$s), 3) calculating the reward (2.1 $\mu$s), and 4) updating the Q-tables (0.5 $\mu$s). The total memory requirement of FedGPO is also efficient --- in our experiments with three device categories, 0.4MB, 0.0125\% of the 32GB DRAM capacity, was only required.

%\vspace{-0.2cm}
\section{Related Work}
\label{sec:related work}

\begin{comment}
\textbf{Energy optimization for mobile devices:} 
Given the stochastic nature of the mobile execution environment, several dynamic energy management approaches have been proposed for individual devices capturing the uncertainties with statistical models~\cite{BGaudette2016}. Other computation offloading techniques have been also proposed for energy efficiency optimization considering the performance variability inherent to the mobile environment~\cite{TAlfakih20,MAltamimi2015,ECuervo2010,LHuang2020,YGKim2017_2,YGKim2019,SPan2019,LQuan2018,BZhang2020,TZhang2019,YGKim2020}.
Although the aforementioned techniques address similar runtime variance to that in the edge-cloud execution environment, they are not appropriate for FL optimization, as they do not consider the unique characteristics of FL use cases --- along with the runtime variance, system and data heterogeneity can induce additional uncertainties in FL execution efficiency.
\end{comment}

\textbf{Optimization for FL:} FL enables collaborative training of a shared ML model in a private manner~\cite{KBonawitz2019,JKonecny2016,HBMcMahan2017,SLin2020}. To deploy FL on the edge efficiently, FedAvg has been employed as the standard algorithm~\cite{JKonecny2016,HBMcMahan2017}. This algorithm alleviates the communication overheads by employing fewer participants with higher number of per-device training iterations~\cite{HBMcMahan2017,YDeng2021,CChen2021}. Subsequently, various methods have been proposed for model accuracy improvement~\cite{TLi2020_2} and security robustness~\cite{YLi2018,YLu2020}.

Although FedAvg demonstrates the potential of FL deployment, it is still faceing the important challenges for optimization --- namely the straggler problem and data heterogeneity. As the straggler problem leads to the excess energy consumption of participant devices, prior studies attempted to mitigate it via asynchronous gradient updates~\cite{YChen2019} or intelligent participant selection~\cite{YJin2020,HBMcMahan2017,YGKim2021}. In addition, to deal with the adverse impact of data heterogeneity, other approaches attempted to reduce the amount of non-IID data reflected on model gradients by allowing asynchronous aggregations~\cite{YChen2019,YChen2019_2}, the use of globally shared data~\cite{YZhao2018}, or partial updates~\cite{TLi2020}. As the main target of the techniques does not encompass the global parameter adjustment, they can be applied with FedGPO.

\textbf{Global parameter optimization for FL:} Along with the general optimizations, global parameter optimization is also crucial in FL, as the global parameters can significantly affect the FL execution efficiency. 

In traditional centralized training, many hyperparameter optimization (HPO) techniques have been proposed to expedite the space exploration based on ML. Such techniques include Bayesian Optimization (BO)~\cite{ASouza2020,FHutter2019} and the Genetic Algorithm (GA)~\cite{HAlibarahim2021}. However, as these HPO techniques require training with the entire dataset for each set of global parameters, they are not feasible for the resource-constrained edge execution environment. Furthermore, they also do not directly address the unique challenges of FL including the straggler problem and data heterogeneity.

Considering the system and data heterogeneity in FL, several approaches have been proposed to adjust the global parameters round-by-round based on exponentiated gradient updates~\cite{MKhodak2021} or deep RL~\cite{ZMa2021}. However, these techniques do not consider the stochastic nature of the edge-cloud execution environment including performance interference and network variability. In addition, they do not account for the energy efficiency of the participant devices. Different from the prior approaches, FedGPO explores the energy efficient global parameter optimization for FL, when there exists system/data heterogeneity and runtime variance. Based on a customized reinforcement learning, FedGPO can identify a near-optimal global parameters for each round by adapting to the heterogeneity and runtime variance. 

\vspace{-0.2cm}
\section{Conclusion}
\label{sec:conclusion}

To enable energy-efficient FL on the edge, we propose a global parameter optimization framework called FedGPO. The FL performance and energy efficiency characterization in the edge execution environment demonstrates that optimal global parameters depend on various features: NN characteristics, system/data heterogeneity, and stochastic runtime variance. FedGPO successfully determines near-optimal global parameters in consecutive FL aggregation rounds, by considering these features. We implement representative FL use cases on an emulated edge-cloud execution environment using off-the-shelf systems. FedGPO improves the average FL energy efficiency by 3.6x, compared with the baseline settings. Compared to FedEX and ABS, FedGPO improves the energy efficiency by 1.5x and 2.1x, on average, respectively, by considering system/data heterogeneity and runtime variance. We demonstrate the viability of FedGPO as a solution to global parameter optimization for energy-efficient FL in realistic edge-cloud execution environments.  

%\section*{Acknowledgments}
%This work is supported in part by the National Science Foundation under grants CCF-1652132 and CCF-1618039 and by the National Research Foundation of Korea under grant NRF-2021R1C1C1008617 and MSIT (Ministry of Science and ICT) of Korea under the ICT Creative Consilience Program (IITP-2022-2020-0-01819) supervised by the IITP (Institute for Information & Communications Technology Planning & Evaluation) for ASU and Korea University. 

\bibliographystyle{IEEEtranS}
\bibliography{IEEEexample}

\end{document}